\documentclass[letterpaper, 10 pt, conference]{ieeeconf}  

\IEEEoverridecommandlockouts                              
\usepackage{graphicx}
\usepackage{color,soul}
\usepackage{hyperref}
\usepackage{mwe}
\usepackage{subcaption}
\usepackage{amsmath}
\usepackage{tabularx} 
\usepackage{multirow}

\newcolumntype{L}{>{\centering\arraybackslash}m{3cm}}
\usepackage{float}
\usepackage[thinc]{esdiff}
\usepackage{graphicx} 
\usepackage[table]{xcolor}
\usepackage{caption}
\title{\LARGE \bf
SLURP! Spectroscopy of Liquids Using Robot Pre-Touch Sensing
}

\author{Nathaniel Hanson$^{1*}$, Wesley Lewis$^{2*}$, Kavya Puthuveetil$^{2*}$, Donelle Furline$^{1}$, Akhil Padmanabha$^{2}$, \\Taşkın Padır$^{1}$, Zackory Erickson$^{2}$ 
\thanks{$^{*}$These authors contributed equally.}%
\thanks{This research is supported by the National Science Foundation under Award
Number 1928654 and National Science Foundation Graduate Research Fellowship Program under Grant No. DGE1745016 and DGE2140739.}
\thanks{$^{1}$Institute for Experiential Robotics, Northeastern University, Boston, Massachusetts, USA.}%
\thanks{$^{2}$Robotics Institute, Carnegie Mellon University, Pittsburgh, Pennsylvania, USA.}%
\thanks{Corresponding author: \tt\small hanson.n@northeastern.edu}%
\thanks{Project Code, Materials List, and Assembly Instructions: \url{https://github.com/RIVeR-Lab/slurp_grasping}
}
}

\begin{document}

\maketitle
\thispagestyle{empty}
\pagestyle{empty}

\begin{abstract}
Liquids and granular media are pervasive throughout human environments. Their free-flowing nature causes people to constrain them into containers. We do so with thousands of different types of containers made out of different materials with varying sizes, shapes, and colors. In this work, we present a state-of-the-art sensing technique for robots to perceive what liquid is inside of an unknown container. We do so by integrating Visible to Near Infrared (VNIR) reflectance spectroscopy into a robot's end effector. We introduce a hierarchical model for inferring the material classes of both containers and internal contents given spectral measurements from two integrated spectrometers. To train these inference models, we capture and open source a dataset of spectral measurements from over 180 different combinations of containers and liquids. Our technique demonstrates over 85\% accuracy in identifying 13 different liquids and granular media contained within 13 different containers. The sensitivity of our spectral readings allow our model to also identify the material composition of the containers themselves with 96\% accuracy. Overall, VNIR spectroscopy presents a promising method to give household robots a general-purpose ability to infer the liquids inside of containers, without needing to open or manipulate the containers.

\end{abstract}

\section{INTRODUCTION}
\begin{figure}[t]
    \vspace{0.5em}
    \includegraphics[width=\linewidth]{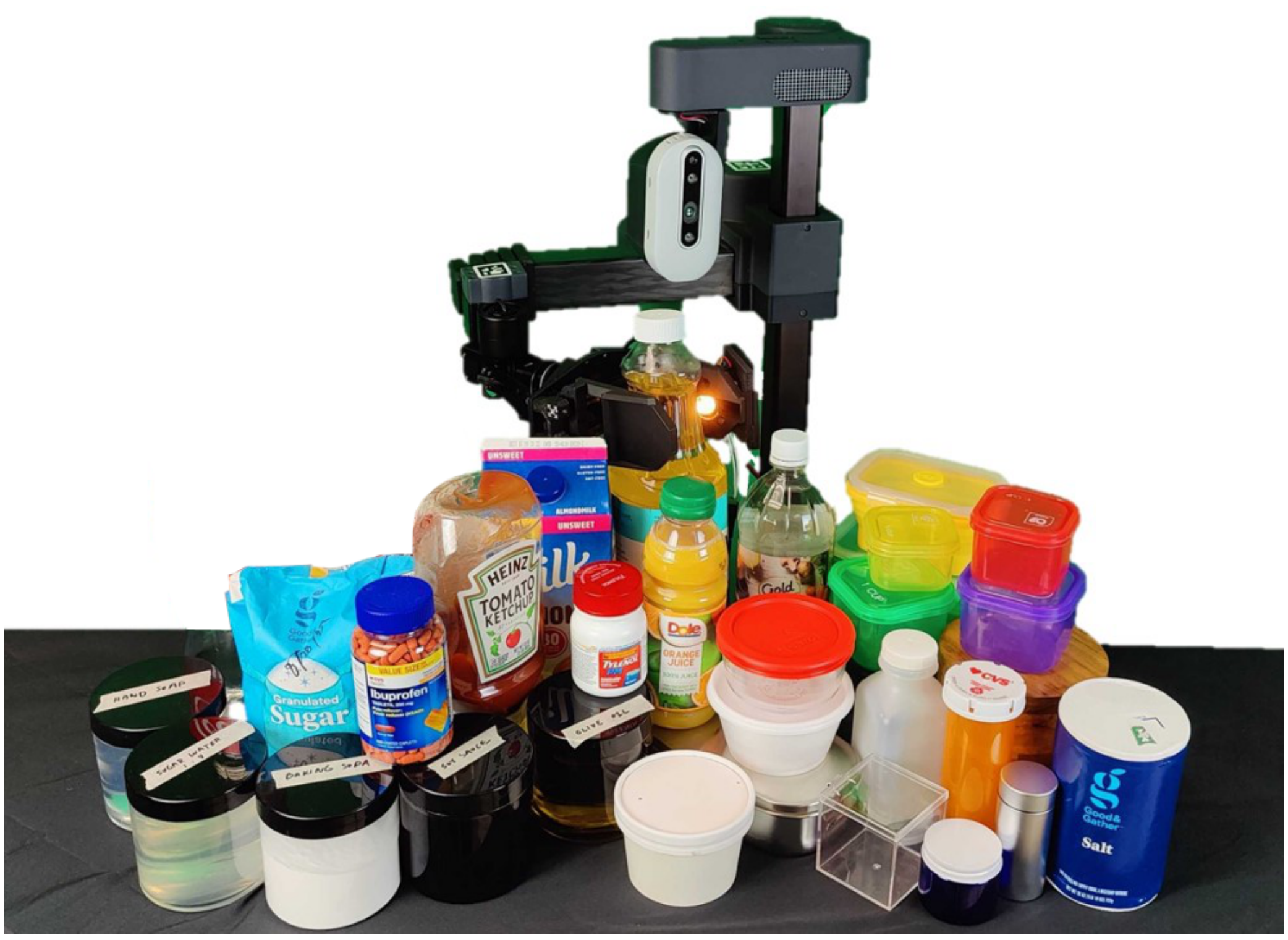}
    \caption{SLURP gripper mounted on a mobile manipulator (Stretch RE1), surrounded by a variety of containers and contents that the novel system can identify without touch.}
    \label{fig:slurp_spread}
    \vspace{-2.0em}
\end{figure}
Knowing liquid properties is a critical first step in properly handling both open and closed containers.  Manipulating these containers is a general-purpose skill robots must master to function in a domestic environment, as everything from medicine to food is stored inside containers. Containers give defined shape to their otherwise formless contents, but liquids' variable viscosities and movable centers of gravity motivates a need to understand what a container holds before it is manipulated.  Visual sensors (cameras and LIDARs) are challenged by these scenarios as container opacity limits vision of the internal contents. Tactile sensing is also constrained by the diversity of liquid densities. Opening a container with a manipulator to directly observe the contents is a challenging dexterous manipulation task, and would still require sensing the internal liquid. The core question of this research is: \textit{Can robots uniquely and reliably identify liquids without opening or manipulating their containers?}

Our solution to this problem leverages spectroscopy to see into containers and understand their contents. Spectroscopy is a technique adapted from the domain of analytical chemistry used to determine the composition of objects through light reflected off their surface. The proportions of light reflected, transmitted, and absorbed at different wavelengths can be used to fingerprint a material's properties and composition. This paper introduces a new sensing platform for inferring the unknown liquid contents inside an unknown container. To accomplish this, we integrate micro near-infrared spectrometers into a robot manipulator, enabling it to optically sense liquids inside visually opaque containers without opening or otherwise manipulating the container.

In this research, we focus on the design, development, and validation of VNIR spectroscopy to perceive liquids and granular solids in containers, shown in Fig~\ref{fig:slurp_spread}. Our work addresses the fundamental question of whether a liquid can be identified in its container through spectroscopic measurements.  Our results demonstrate spectroscopy is a powerful technique that can be used to recognize a variety of containers and their contents with minimal training data.

The contributions of this paper are:
\begin{itemize}
    \item Development of a gripper system to obtain VNIR spectral signatures from solids and liquids inside containers.
    \item Dissemination of a dataset of complex container-content spectral signatures.
    \item Demonstration of signal processing and a hierarchical neural network to generate container and content predictions at grasp time.
\end{itemize}

\section{RELATED WORK}
\label{sec:prior}

\subsection{Robot Liquid Perception}


Liquid identification remains an open and challenging problem in robotics. Prior work has demonstrated good results in identifying containers by their shape as a proxy to identifying the internal substance \cite{scavino2009application,sakr2016comparing}. \cite{do2016probabilistic} used an RGB-D camera to segment liquid level and measured refraction to predict fill contents. Others have presented methods of identifying liquid or granular media in containers by measuring the response of the contents when the container is perturbed by an applied force \cite{matl2019haptic, huang2022understanding, saal2010active, guler2014s}. The utility of these tactile estimation methods, however, is dependent on the substrates having distinct viscosity (e.g. water vs. yogurt) and classification results for substances with similar viscosity remains unknown.

The other core research thrust in liquid perception is the modeling of poured liquids. Segmenting semi-transparent liquid flows \cite{yamaguchi2016stereo,narasimhan2022self} has been studied to enable better control over liquids streams between containers. However, this scenario presupposes that the liquid is in an open container, and that pouring the liquid from one container to another is desired and will not create an undue mess.

\subsection{Liquid Spectroscopy}


In traditional imaging, red, blue, and green channels sense reflected light at three central wavelengths. Spectroscopy increases both the number and specificity of the sensed wavelengths of light. The near infrared region (800 - 2500 nm) is noted for the particular presence of anharmonic vibrational modes where NIR light is absorbed and reflected in distinct quantities as a function of the abundance of chemical bonds \cite{pasquini2018near}. Within the domain of NIR spectroscopy, transmission spectroscopy is typically used to analyze the chemical composition of samples by measuring the attenuation of light as it passes through a small liquid sample \cite{blanco2002nir}. The efficacy of spectroscopic techniques has been demonstrated in predicting fat and lactose contents in dairy milk
\cite{aernouts2011visible}, as well as in estimating the freshness \cite{aboonajmi2014prediction}, composition \cite{genisheva2018new}, and authenticity \cite{chen2014discrimination} of various foods. \cite{wang2017quality} provides an excellent review of the breadth of liquid foods successfully analyzed by NIR spectroscopy.

\subsection{Robot Spectroscopy}

The use of spectroscopy in robotics is an emerging field and has been demonstrated, often in combination with additional sensors, in a number of contexts including: classification of household objects \cite{erickson2019classification, erickson2020multimodal}, ripeness evaluation of mango fruit \cite{cortes2017integration}, in-hand object recognition \cite{Hanson2022-Flexible}, and ground terrain characterization \cite{Hanson2022_VAST}.

These research methods have all yielded highly accurate classification results on substances with distinct material compositions. This current research is distinct from prior work since it proposes spectroscopy as a useful method to classify visually transparent materials. Previous research focused on sensor design for opaque surfaces with no internal contents. Our work adds emphasis on recognizing the internal contents of the container. To the authors' best knowledge, this research is the first to propose a non-contact method for robots to estimate the internal contents of containers.

\section{MATERIALS \& METHODS}
\label{sec:methods}
Unlike surface reflectance spectroscopy, which assumes an opaque object surface, spectroscopy of liquids must also work with a range of container transparencies. To this end, we created a specialized gripper to provide active illumination in the VNIR range, under which containers exhibit transparency \cite{stark2015short}. We provide parts listings, models, and assembly documentation in the project repository.
\begin{figure}[t]
    \vspace{0.5em}
    \includegraphics[width=\linewidth]{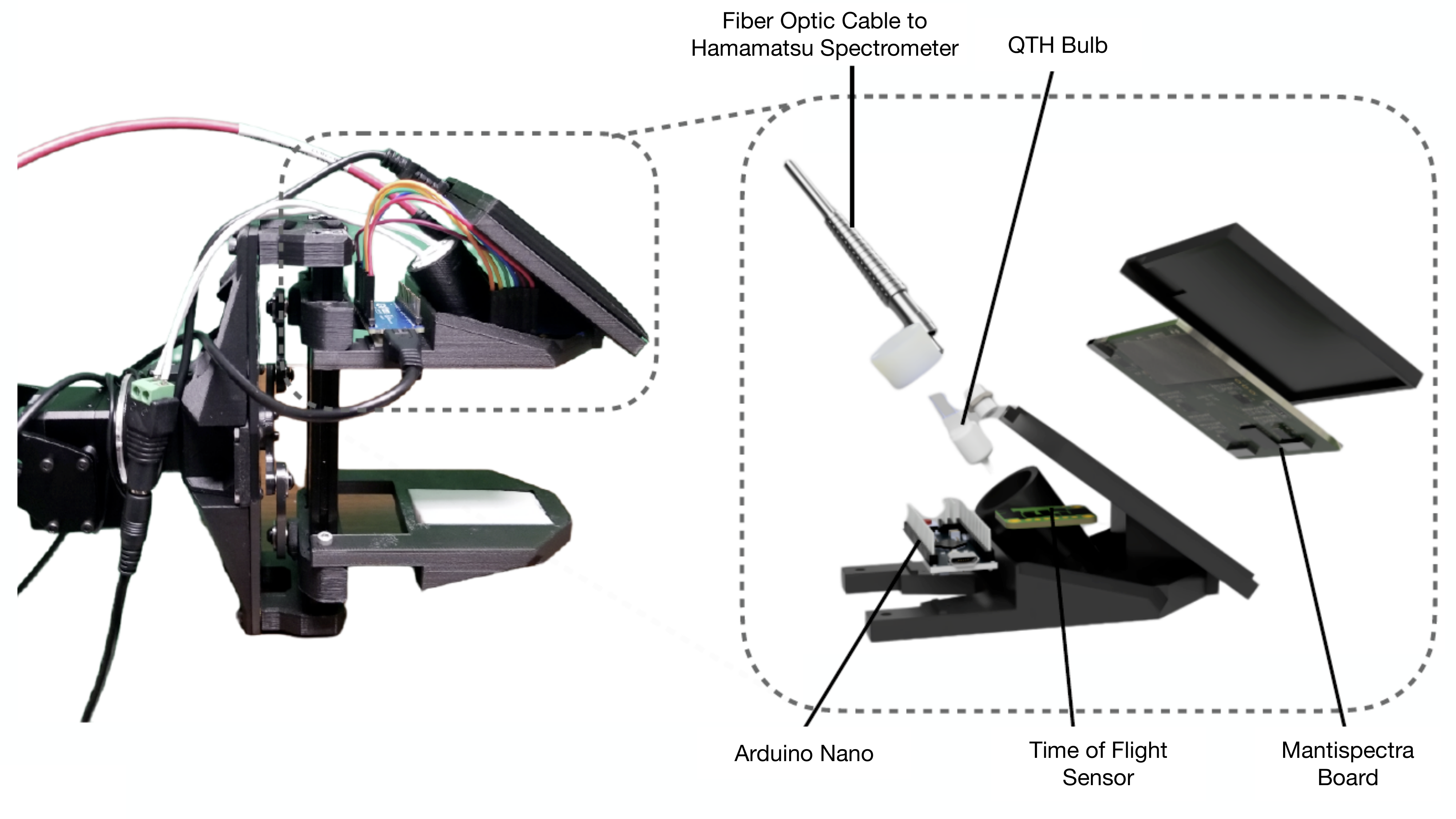} 
    \caption{(Left) Real-world assembly of the SLURP gripper. (Right) Exploded CAD rendering of SLURP gripper paddle showing integrated visible to near infrared spectrometers and active illumination associated physical gripper assembly.}
    \label{fig:exploded_gripper_render}
     \vspace{-1.5em}
\end{figure}

\subsection{Optical System Design}
 We developed a parallel jaw gripper with a linear driver (building on the ViperX 300 gripper from Trossen Robotics) with integrated spectroscopy sensing. Fig. 2 depicts this gripper and the key sensing components. We 3D printed the paddles in Onyx \cite{onyx2022} for added rigidity and thermal stability. The dimensions of the modified sensor paddle are 106 mm $\times$ 69 mm $\times$ 51 mm with a contact surface area of $\approx$48 $\text{cm}^2$. The large surface area provided requisite space to mount sensors. Our system makes use of both paddles in our signal acquisition procedure. First, light from a Quartz Tungsten Halogen (QTH) bulb provides active illumination onto the nearby container surface. The specific bulb (ThorLabs QTH10B) was selected since it provides even illumination in the 400-2500 nm range. The bulb was operated at 6v and exhibited an average surface temperature of $36^{\circ}$ C --- well within the safe working temperature of common containers. The bulb is recessed from the surface plane of the paddle to avoid direct contact with grasped items. A coated reflector helps focus the illumination on the target item.

In contrast to our previous research which used single sensors with primary sensitivity in the visible range, this work incorporates two spectrometers into a gripper to extend the sensed wavelength information. This redesign is necessary to account for spectral variability caused by the container. To capture the visible range, we utilize a silicon-detector, micro-electro-mechanical system (MEMS) spectrometer (Hamamatsu C12880ma)\footnote{\url{https://www.hamamatsu.com/jp/en/product/photometry-systems/mini-spectrometer}}. This device collects photons in 288 distinct spectral channels, with an average spectral resolution of 9 nm from 340-850 nm. We avoid additional bulk on the gripper by utilizing a Low-OH (hydroxyl group) fiber optic patch channel. The cable is mounted to an SMA connector embedded in the gripper paddle. These cables have a high numerical aperture (NA) to maximize the transmitted photons. The other end of the fiber optic cable terminates to a mounting plate centered over the input slit of the spectrometer.

Traditional NIR spectrometers are an order of magnitude more expensive due to the Indium Gallium Arsenide photodetector needed for perception beyond 1000 nm. To minimize costs, we incorporate a spectrometer-on-chip design (Mantispectra OEM Board)\footnote{\url{https://www.mantispectra.com}} with an operating range of 850-1700 nm. Unlike a traditional grating-based spectrometer, where a machined diffraction grating aligns light with a detector cell, the channels on this device read a non-linear signal consisting of a bi-modal response specific to each channel. Hakkel et al.~\cite{hakkel2022integrated} provide additional details on the design of this spectrometer. Rather than unmixing the response of each channel into its component wavelengths~\cite{zhang2018spectral}, we use the 16-dimensional signal from this spectrometer in its original form, consisting of 16 channels of measured photocurrent. Combined, this new optical acquisition system enables a robot to capture NIR spectral signatures across the full VNIR range of 340-1700 nm.

Finally, the gripper also contains an integrated time-of-flight (ToF) infrared distance sensor to precisely measure the distance from the gripper paddle to in-hand objects. A small standoff distance, $\approx$1 cm, between the sensor and object surface is needed to ensure adequate collection of photons into the sensors. Similarly, too great a standoff distance can allow ambient reflections into the spectrometer, thus skewing the acquired signal \cite{weatherall2012feasibility}. The device communicates via serial commands with a small embedded controller (Arduino). All the systems are controlled with a Python interface compatible with the Robot Operating System (ROS)~\cite{quigley2009ros}. An exploded view rendering of the gripper is depicted in Fig~\ref{fig:exploded_gripper_render}.

\begin{figure}[t]
    \vspace{0.5em}
    \includegraphics[width=\linewidth]{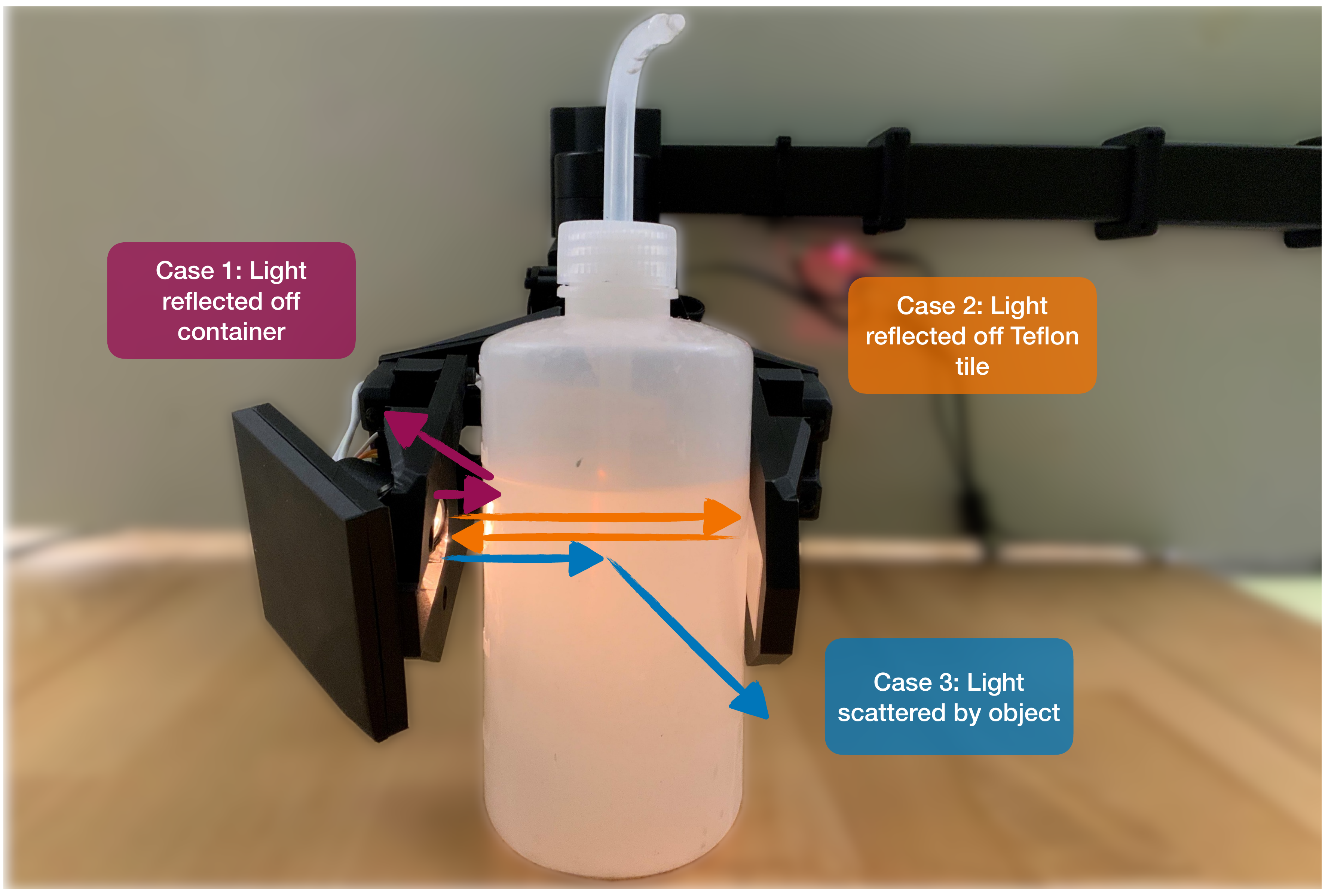}
    \caption{Light interactions with containers and internal contents from photon creation to sensor measurement.}
    \label{fig:signal_acquisition}
     \vspace{-2.0em}
\end{figure}

\subsection{Operating Principle}
The signal acquisition process is depicted in Fig~\ref{fig:signal_acquisition}. Photons from the light source are transmitted through a target container within the gripper. They then interact with both the container and the liquid inside. Some photons are scattered or absorbed by the contents; others are reflected back to the spectrometer after interacting with the container and its contents. 
\begin{figure*}[t!]
    \vspace{0.5em}
    \includegraphics[width=\linewidth]{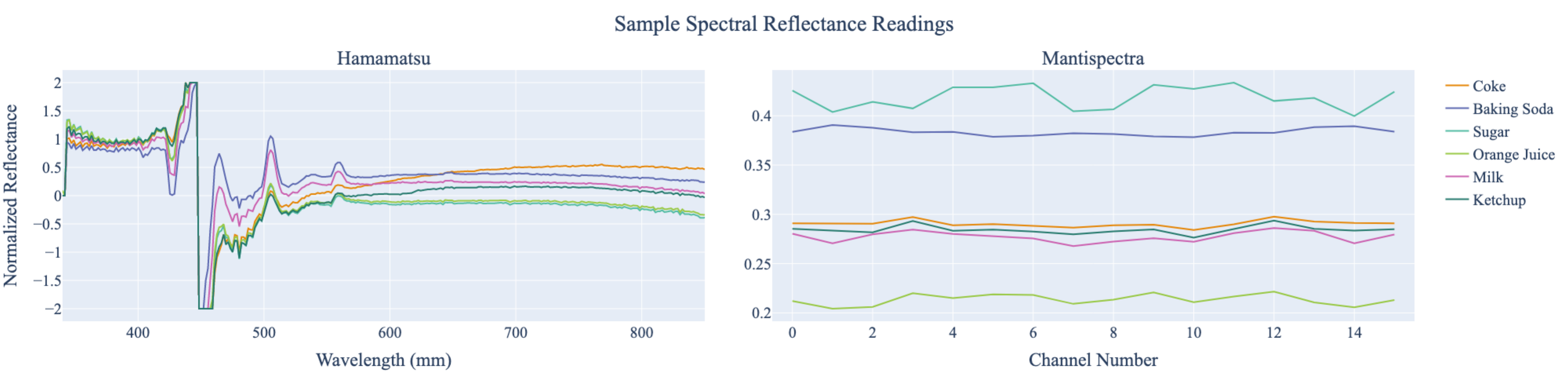}
    \caption{Sample calibrated spectral readings as viewed through a quartz cuvette (optically transparent). Note, the split axes show the normalized readings as a function of wavelength for the Hamamatsu spectrometer and detector channel for the Mantispectra chip.}
    \label{fig:sample_spectra}
     \vspace{-1.5em}
\end{figure*}
For highly transmissive materials, such as water, we maximize the signal returned to the spectrometer by embedding a sheet of white Teflon into the opposite gripper. This material is a cost-effective approximation of a Spectralon white reference standard as it also reflects $>95\%$ of incident light across the VNIR spectrum. Light reflected off this surface then travels back through the container and substrate where further interactions occur. The combination of light lost from scattering, reflectance, and absorption all contribute to the formation of spectral readings, consisting of intensity of light, as a function of wavelength, that makes it back to the spectrometer apertures.

\section{DATASET}
\label{sec:dataset}
We assemble a large dataset of containers and internal substrates to explore the breadth of material detection and classification accessible by spectroscopy. For containers, we selected common household objects that varied in material, optical clarity, geometry, thickness, and color. With respect to optical clarity, we include a variety of containers that range from translucent to near opaque. This inclusion is grounded in previous findings that certain plastics exhibit transparency under NIR lighting, while liquids become highly reflective~\cite{stark2015short}. The substrates we selected include common pantry items and medicines that are difficult to distinguish via RGB vision alone (e.g. sugar vs. salt). In addition to liquid substrates, we chose to include several granular materials in our data collection since they are amorphous solids routinely stored inside containers.

\subsection{Data Collection Process}
To collect a large, diverse dataset consisting of spectral measurements from a variety of liquids inside many different containers, we mounted the sensorized gripper paddles onto a stationary rail to allow for rapid data collection. Next, we 3-D printed a Polylactic acid (PLA) spacer to allow consistent offset spacing between each container and the sensor array (1 cm). This jig helps accelerate data collection and is not used during real robot operation (Section~\ref{sec:robot}). Spectrometers collect data over an interval known as integration time. Each sensor has an intrinsic digitization level, i.e., the point at which its photodetector levels saturate. Moreover, the operating voltage of the active illumination source controls the upper limit of available photons to sense. As each of the included spectrometers operates independently, we tuned their integration times to allow for the maximum possible signal strength across the spectrum, while avoiding sensor saturation. The Hamamatsu used an integration time of 170 ms; the Mantispectra device used 512 ms; the QTH light was operated at 6V. The small (1 cm) air gap and focused illumination ensure the signal to the spectrometer is dominated by QTH incident spectrum with minimal environmental light.


Our dataset includes spectral measurements with the following container types and colors: Glass (\textit{blue, clear}); Polypropylene PP (\textit{yellow, green, red, orange, and purple}); Polyethylene Terephthalate PET (\textit{clear}); Paper (\textit{white}); Acrylic PMMA (\textit{clear}); Silicone (\textit{yellow, green, and blue}). 



Each of the containers was successively filled with the following 13 content classes: \textit{Almond Milk, Vegetable Oil, Olive Oil, Water, Ibuprofen, Acetaminophen, Orange Juice, Ketchup, Soy Sauce, Coke, Salt, Sugar, and Empty}.

Fig~\ref{fig:sample_spectra} shows sample spectra through a quartz container---a commonly used scientific standard for measuring transmission spectra of liquids. We captured 30 samples using both spectrometers for each suitable container and substrate combination. This resulted in a total dataset of 5,070 spectral measurements for each spectrometer across all $13 \times 13$ container-substrate combinations (including an ``empty'' content class). After each consecutive measurement, we adjusted the orientation and position of the container inside the gripper. By translating the container up to 20 mm towards the gripper's tip or rail mounting, we capture a distribution of spectral samples along a container's surface before rotating the container clockwise to collect more samples. 


\subsection{Signal Processing}
\label{sec:sig_processing}
After collecting raw samples from both the spectrometers, the samples are normalized using the following equation adopted from \cite{geladi2004hyperspectral}:
\begin{equation}
S_{cal spec} = \frac{S_{spec} - {\tilde{D}}_{spec}}{{\tilde{L}}_{spec} - {\tilde{D}}_{spec}}
\end{equation}
In this formulation, the tilde operator refers to the median of the class samples. Median is preferred to mean as it is more robust to noise saturating pixels. The median measurements are taken over ten consecutive readings for each reference measurement. ${\tilde{L}}_{spec}$ refers to the median measurement taken from the white Teflon block integrated into the opposite gripper from the 1 cm air-gapped standoff. Signals collected from the Teflon block in the opposite paddle with no object in the gripper represent the maximum possible signal at the given distance. ${\tilde{D}}_{spec}$ refers to the median measurement taken in a dark room with nothing directly in front of the spectrometer. This reading accounts for dark current, and other electronic noise when the sensor is not actively detecting photons. ${S_{spec}}$ refers to the spectral signal for a particular sample and ${S_{calspec}}$ is the calibrated output.

\section{MODELING}
\label{sec:modeling}
As a practical demonstration of the utility of spectral signatures to see into containers, we develop a hierarchical modeling technique to contextualize the spectral reading into container and contents. 
From a modeling perspective, we explore two inference questions:
\begin{enumerate}
    \item What is the container composed of?
    \item What does the container hold inside?
\end{enumerate}
\subsection{Hierarchical Modeling}
We approach the problem of modeling incoming spectral readings as a mixture problem. Namely, as light penetrates the surfaces of the container and illuminates the contents, some incident wavelengths will exhibit a loss of intensity as they are absorbed. Consequently, any readings at that particular wavelength of the substance in the container will also be diminished. A na\"ive approach to this problem would attempt to learn separate models to predict container and content classes. In contrast, our hierarchical approach has the added benefit of carrying estimates of the container type into the estimation of the contents.

Formally, we define the container decision problem as:
\begin{align}
    \label{eqn:formalism_level_one}
    \hat{x} = \textit{argmax} (p_{\theta}(x|z))
\end{align}
where $\hat{x}$ is the predicted container class, $\theta$ is the parameters for a fit model, and $z$ is the current observation from the spectrometer. We then predict a class label $\hat{\beta}$ for the internal substrate (liquid or granular solid) conditioned on both the spectral observation $z$ and the predicted container class $\hat{x}$. 
\begin{align}
    \label{eqn:formalism_level_two}
    \hat{\beta} = \textit{argmax} (p_{\theta}(\beta|z,\hat{x})).
\end{align}

\subsection{Implementation \& Training}

\begin{figure}
    \vspace{0.5em}
    \includegraphics[width=\linewidth]{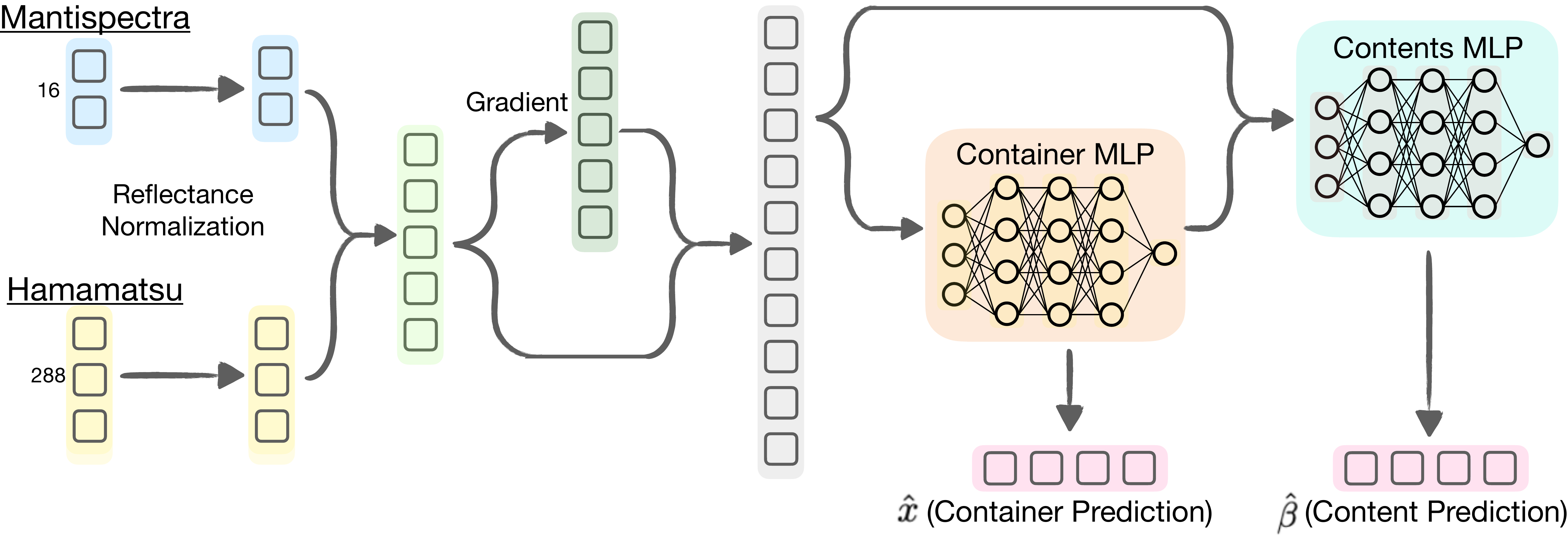}
    \caption{Network architecture to learn simultaneous inference of container and content material.}
    \label{fig:network_architecture}
     \vspace{-1.5em}
\end{figure}
We implement models using multi-layer perceptrons (MLP) as approximators for these conditional probabilities. The detailed network architecture is depicted in Fig~\ref{fig:network_architecture}. We learn both the container and content models simultaneously, with the current predictions of the container concatenated with the original spectral signature. Each MLP layer consists of 3 fully connected layers of 200, 100, and 25 neurons, respectively. Each layer is followed by a rectified linear unit (ReLU) activation function and a dropout layer of probability of 0.25 to mitigate overfitting. The input vector to the model is the reflectance normalized spectral reading concatenated to the gradient of the signal ($z \cup \nabla z$). For a vector, $\nabla z$ is generated by using finite differencing. The returned gradient hence has the same shape as the input array making each input reading a 608 length vector.

We compute two cross entropy losses for our network; $\mathcal{L}_\text{container}$ for the container MLP branch, and $\mathcal{L}_\text{content}$ for the (liquid prediction) content MLP. Both losses take the form:
\begin{align*}
    \mathcal{L} = -\sum^{N}_{c=1}y_{z,c}\log({p_{z,c}})
\end{align*}
In this definition, $N$ is the number of classes predicted, $y_{z,c}$ is the a one-hot vector denoting class membership; and $p_{z,c}$ is the network prediction of the class likelihood, given a measurement, $z$. The composite loss function is created by summing the two constituent network losses.
\begin{align*}
    \mathcal{L}_\text{total} = \mathcal{L}_\text{container} + \mathcal{L}_\text{content}
\end{align*}
This composite loss is back-propagated through both networks. Before training, we augmented the existing data by a factor of three by duplicating the normalized data and randomly applying a multiplicative factor in range $[0.8,1.2]$ to the data copies. This approach augments our dataset size and accounts for minor variations in signal strength due to changes in ambient environmental illumination or slight specular reflections. Data was split into train, validation, and held-out test samples in an 80/10/10\% ratio. The test and validation sets were stratified against the labels to ensure appropriate class balance. The models were implemented in PyTorch \cite{paszke2019pytorch} using GPU acceleration on an NVIDIA GeForce RTX 3080. We trained models for 500 epochs using the Adam Optimizer~\cite{kingma2014adam} with a batch size of 64 and an initial learning rate of 0.01. We trained ten consecutive models on our dataset, keeping the model that scored highest on the held-out test data
to deploy on the robot.

\section{Results}
\label{sec:results}
\begin{table}[t]
    \vspace{0.5em}
    \centering
    \normalsize
    \begin{tabular}{|| c | c | c | c ||}
        \hline
        Model & Container Acc & Content Acc & Acc\\
        \hline \hline
        Hierarchical & \textbf{0.96} & \textbf{0.85} & \textbf{0.82}\\
        \hline
        Hamamatsu & 0.92 &  0.78 & 0.73\\
        \hline
        Mantispectra & 0.93 & 0.57 & 0.54\\
        \hline
        No Gradient & 0.94 & 0.79 & 0.76\\
        \hline
        No Augment & 0.95 & 0.81 & 0.78\\
        \hline
        Na\"ive Models & 0.88 & 0.72 & -\\
        \hline
    \end{tabular}
    \caption{Model results evaluated on the held-out test dataset averaged over ten random seeds, including ablated results where models are trained with a smaller input feature vector.}
    \label{tab:results}
    \vspace{-1.0em}
\end{table}

\begin{table}[t]
    \vspace{0.5em}
    \centering
    \normalsize
    \begin{tabular}{|| c | c | c | c | c | c ||}
        \hline
        Acrylic & Glass & PET & PP & Paper & Silicone\\
        \hline \hline
        0.97 & 0.99 & 0.96 & 1.0 & 1.0 & 1.0\\
        \hline
    \end{tabular}
    \caption{Overall accuracy for each of the container types in the container model network predicted using the best set of network weights on test data.}
    \label{tab:container_results}
    \vspace{-2.0em}
\end{table}
The results of the model predictions on the held-out test dataset are detailed in Table~\ref{tab:results}. In this table, we analyze three important predictive qualities of the model: container accuracy is accuracy of the output container predictions ($\hat{x})$, content accuracy highlights the accuracy of the predicted liquid contents ($\hat{\beta}$), and overall accuracy (Acc) shows performance of the model to distinguish both quantities correctly.

The proposed hierarchical method outperforms ablated versions of the model. Compared to the na\"ive implementation with two completely disjoint networks, the hierarchical technique is 8\% more accurate on containers, and a 13\% improvement on contents. Because the na\"ive approach trains the two networks independently, the joint overall accuracy is not assessed as it is for all other model types.

We also observe the hierarchical model predicts containers on average 14\% more accurately than the contents. This result implies that the container contributes more to the measured signal that the internal components. This intuitively makes sense, as in the best case, the container will transmit all of a particular wavelength of light, and in the worst case inhibit all of it. Table~\ref{tab:container_results} further breaks down the container accuracy score by container types. These results are derived from the best model trained and validated over 10 random seeds. These results include all intraclass color and opacity variation for the given container. Our model achieved classification accuracies greater than 96\% for all container types. Over multiple seeds, the model occasionally misclassified PET as PP or Acrylic. However, these results were rare and further demonstrate the model's strong performance at estimating external container material regardless of geometry, color, opacity, or content.

\begin{figure}[t]
    \vspace{0.5em}
    \includegraphics[width=\linewidth]{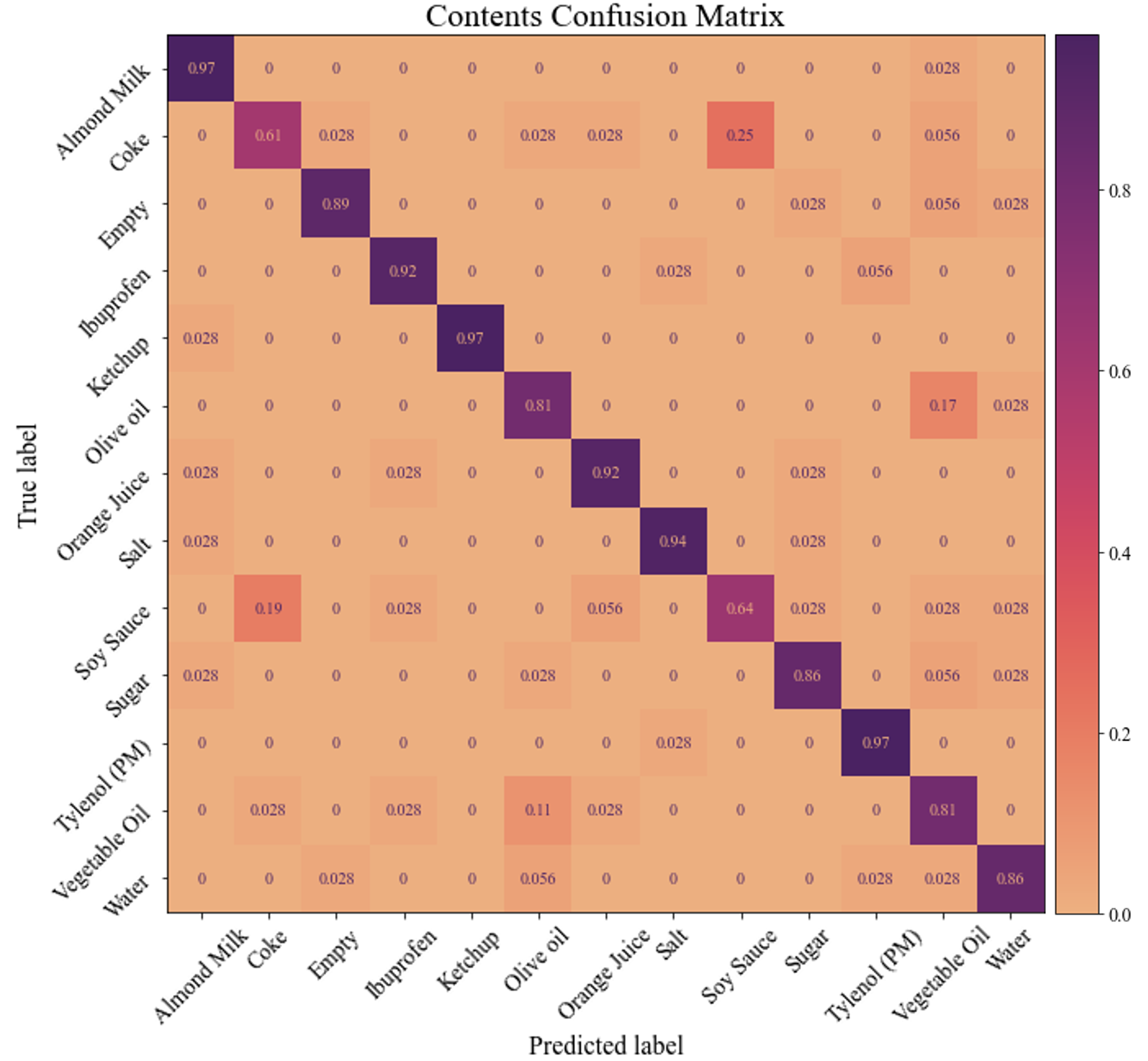}
    \caption{Confusion matrix for model predictions of contents across all container combinations in the held-out test dataset.}
    \label{fig:contents_confusion}
    \vspace{-1.5em}
\end{figure}

Fig~\ref{fig:contents_confusion} presents a confusion matrix for our model's classification performance on the held-out test set across all liquid content types. Overall, our model correctly classified most liquids with accuracies over 80\% and often over 90\% regardless of the container that the liquid was held in. Two notable exceptions were soy sauce and coke. These substances are both optically darker, generally leading to lower signal returns. They are also are fully dissolved liquids, unlike colloids of suspended particles (e.g. ketchup and almond milk) which have larger molecular structures and provide stronger readings. 

As an ablation to the the full model architecture, we also train the same network using only a single spectrometer. The overall lower accuracy of the Hamamatsu-only (73\%) and Mantispectra-only (54\%) networks highlight the value of including all available wavelength information. We also ablate our model by removing the calculated gradient from the model. Not including the gradient in the input vector decreases the container and content accuracies by 2\% and 6\%, and the combined accuracy by 6\%. Finally, we remove the random scaling data augmentation. The degraded accuracies show the simulated illumination scaling contributes to the hierarchical model's success.

\section{ROBOT EXPERIMENTS}
\label{sec:robot}

We further evaluate our trained classifier in a real-time experiment using a Stretch RE1 (Hello Robot) mobile manipulator \cite{kemp2022design}. We outfit the mobile manipulator with the 2 DoF wrist components from a ViperX 300 Arm and mount the SLURP gripper to this wrist as shown in Fig.~\ref{fig:exploded_gripper_render}.

\begin{figure}
    \vspace{0.5em}
    \includegraphics[width=\linewidth]{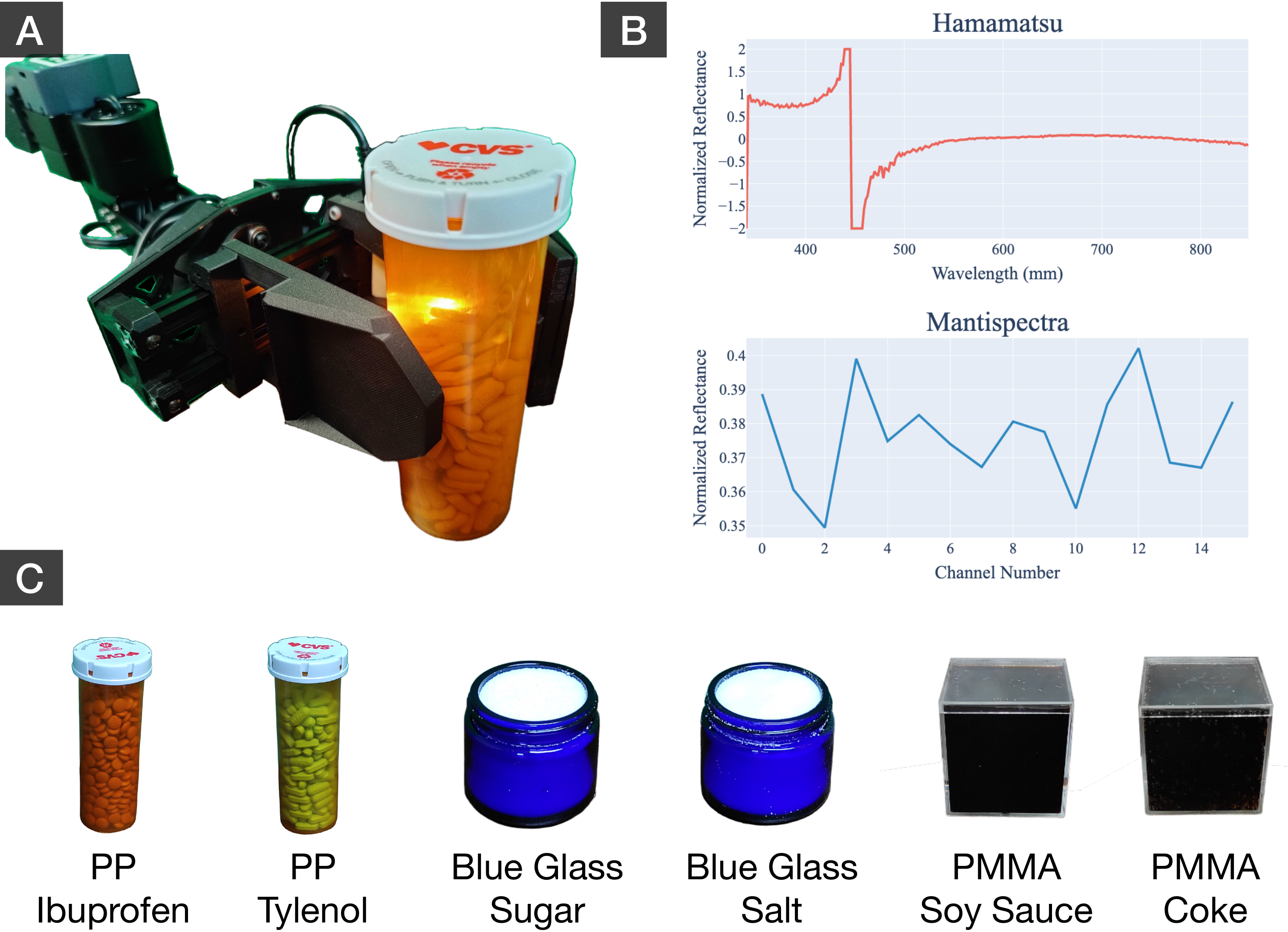}
    \caption{Overview of robot demonstration on tricky scenarios: A) SLURP gripper taking pre-grasp measurements from a PP container filled with ibuprofen, B) Visualization of the spectral readings captured in part A after calibration, C) Examples of container and substrate combinations evaluated in the robot experiment.}
    \label{fig:robots_doing_things}
\end{figure}

We prepare the robot for evaluations by placing a given container on a table in front of the robot's open gripper. The robot closes its gripper around the container until the sensing paddle is 1cm away from the container's outer surface as determined via the SLURP gripper's integrated ToF sensor. The robot then captures a spectral measurement from each spectrometer, processes the signal as detailed in Section \ref{sec:sig_processing}, and feeds the sample into our trained classifier.

We select three containers for the robot experiment (\textit{Polypropylene PP, Glass, and Acrylic PMMA}), all of which fit in the robot's gripper and have varying opacity, geometry, and color. We fill each container with six substrate classes (\textit{Ibuprofen (IBU), Acetaminophen (APAP), Coke, Soy Sauce, Salt, and Sugar}), resulting in a total of 18 different container-substrate combinations. Fig~\ref{fig:robots_doing_things} shows these varied containers and liquid contents. Table~\ref{tab:results_robot} depicts the classification performance of the substrate contents inside of each container.

The overall container and substrate classification accuracies across all tested combinations were 78\% and 72\% respectively. The results in Table~\ref{tab:results_robot} show that ibuprofen, acetaminophen, and soy sauce can be reliably classified between containers. Distinguishing sugar from salt or coke from soy sauce proves to be more difficult, however, this performance is consistent with findings from evaluating our models on the precaptured test set, shown in Fig~\ref{fig:contents_confusion}.


\label{sec:results}
\label{sec:experiments}

\begin{table}
    \centering
    \small
    \begin{tabular}{|| c | c | c | c | c | c | c ||}
        \hline
        Cont & IBU & APAP & Coke & Soy. & Salt & Sugar\\
        \hline \hline
        PP & \cellcolor{green!25}IBU & \cellcolor{green!25}APAP & \cellcolor{red!25}Soy. & \cellcolor{green!25}Soy. & \cellcolor{green!25}Salt & \cellcolor{green!25}Sugar\\
        \hline
        Glass & \cellcolor{green!25}IBU & \cellcolor{green!25}APAP & \cellcolor{red!25}Soy. & \cellcolor{green!25}Soy. & \cellcolor{red!25}APAP & \cellcolor{red!25}IBU\\
        \hline
        PMMA & \cellcolor{green!25}IBU & \cellcolor{green!25}APAP & \cellcolor{green!25}Coke & \cellcolor{green!25}Soy. & \cellcolor{green!25}Salt & \cellcolor{red!25}Salt\\
        \hline
    \end{tabular}
    \normalsize
    \caption{Model substrate predictions from spectral samples collected with the Stretch RE1 from a) containers: PP (\textit{orange}); Glass (\textit{blue}); Acrylic PMMA (\textit{clear}), when filled with b) substrates: \textit{Ibuprofen (IBU), Acetaminophen (APAP), Coke, Soy Sauce, Salt, Sugar}. Successful and failed classifications are highlighted green and red respectively.}
    \label{tab:results_robot}
    \vspace{-1.5em}
\end{table}
\label{sec:results_robot}

\section{CONCLUSIONS}
\label{sec:conclusion}
In this work we presented a first of its kind sensor suite to acquire VNIR spectral signatures from liquids and other granular solids in containers. We detailed an optical sensor design to acquire readings from within a robot gripper. We also released a novel dataset consisting of over 5,070 spectral readings from two complimentary spectrometers taken from 13 different containers filled with 13 different content types. 

We documented the processing and modeling steps to build effective classifiers using spectroscopic data. Furthermore, we demonstrated the gripper's real world effectiveness, operating on a Stretch mobile manipulator to identify containers filled with various liquid and granular solid contents. Overall, VNIR spectroscopy presents a promising method to give household robots a general-purpose ability to estimate both the materials of containers and the contents inside of them, without needing to open or manipulate the containers.



\bibliographystyle{IEEEtran} 
\bibliography{references}

\addtolength{\textheight}{-12cm}   




\end{document}